# A New Repair Operator for Multi-objective Evolutionary Algorithm in Constrained Optimization Problems

Zhun Fan
Guangdong Provincial Key Laboratory of
Digital Signal and Image Processing
Department of Electronic Engineering
Shantou University, Guangdong,
P.R. China
zfan@stu.edu.cn

Wenji Li
Department of Electronic Engineering
Shantou University, Guangdong,
P.R. China
wenji_li@126.com

Xinye Cai
College of Computer Science and
Technology, Nanjing University of
Aeronautics and Astronautics, Jiangsu,
P.R. China
xinye@nuaa.edu.cn

Huibiao Lin
Department of Electronic Engineering
Shantou University, Guangdong,
P.R. China
13hblin@stu.edu.cn

Shuxiang Xie
Department of Electronic Engineering
Shantou University, Guangdong,
P.R. China
12sxxie1@stu.edu.cn

Erik Goodman
BEACON Center for the Study of
Evolution in Action, Michigan State
University, East Lansing, MI, USA
goodman@egr.msu.edu

## ABSTRACT
In this paper, we design a set of multi-objective constrained optimization problems (MCOPs) and propose a new repair operator to address them. The proposed repair operator is used to fix the solutions that violate the box constraints. More specifically, it employs a reversed correction strategy that can effectively avoid the population falling into local optimum. In addition, we integrate the proposed repair operator into two classical multi-objective evolutionary algorithms MOEA/D and NSGA-II. The proposed repair operator is compared with other two kinds of commonly used repair operators on benchmark problems CTPs and MCOPs. The experiment results demonstrate that our proposed approach is very effective in terms of convergence and diversity.

## Categories and Subject Descriptors
G.1.6 [**Optimization**]: Constrained optimization

## General Terms
Algorithms

## 1. INTRODUCTION
Multi-objective optimization problems (MOPs) consist of more than one objectives, which are usually conflicting with each other. In other words, improvements in one objective may lead to the degradation of other objectives. It is impossible to make all of the objectives to be optimal at the same time. Instead, a set of solutions that represent the trade-off between multiple objectives exist for MOPs. In addition, different types of constraints are often unavoidable in MOPs. Such MOPs with constraints are usually termed multi-objective constrained optimization problem. Constraints can be roughly divided into two categories, equality and inequality constraints. Without loss of generality, a multi-objective constrained optimization problem can be defined as follows.

$$\begin{cases} \min & F(x) = (f_1(x), f_2(x), \ldots, f_m(x)); \\ \text{s.t.} & g_i(x) \geq 0, \quad i = 1, 2, \ldots, q \\ & h_j(x) = 0, \quad j = 1, 2, \ldots p \end{cases} \quad (1)$$

Where $x = (x_1, x_2, \ldots, x_n) \subset R^n$ is n-dimensional design variables, $F(x) = (f_1(x), f_2(x), \ldots, f_m(x)) \subset R^m$ is m-dimensional objective vector. $g_i(x) \geq 0$ define $q$ inequality constraints, $h_j(x) = 0$ define $p$ equality constraints.

The existing multi-objective constrained evolutionary algorithms combine the multi-objective evolutionary algorithms with the mechanisms of constraint handling [1]. At present, NSGA-II [2] and MOEA/D [3] are the two classical multi-objective evolutionary algorithms representing two categories of fitness assignment methods, namely fitness assignment based on domination and decomposition. In fitness assignment based on domination, the fitness is decided by non-dominated sorting and crowding distance. Representative algorithms using this type of fitness assignment method include MOGA [4], PAES-II [5], SPEA-II [6] and NSGA-II [2]. In fitness assignment based on decomposition, comparison and sorting of individuals are made via aggregation function with weights allocated specifically to all individuals. Typical algorithms of this category include IMMOGLS [7], UGA [8], cMOGA [9], MOGLS [10], and MOEA/D [3].

The existing constraints handling mechanism can be divided into four categories. They are feasibility maintenance, penalty function, separation of constraint violation and objective value and multi-objective evolutionary algorithms (MOEAs). The methods of feasibility maintenance are usually applied to the discrete optimization problems, such as the job shop scheduling problems and the vehicle routing problems. They either design appropriate coding and decoding methods to ensure that the individuals are feasible, or apply some mechanisms to repair the infeasible individuals. The main idea of penalty function method is adding one penalty term to the objective functions and transforming the constrained optimization problem into an unconstrained one. Typical methods of this category include segregated penalty functions [17], death penalty functions [18], co-evolutionary penalty functions [19] and adaptive penalty functions [20] [21]. The mechanism of separation of constraint violation and objective value treats the objective and constraints separately. Typical methods of this category include stochastic ranking (SR) [11], infeasible driven evolutionary algorithm (IDEA) [12] and constraint dominate principle (CDP) [13]. The main feature of MOEAs is to transform a multi-objective constrained optimization problem to another multi-objective optimization problem with an additional objective, which regards the constraint condition as another objective and uses the existing MOEAs to optimize the transformed problem. Typical methods of this category include COMOGA [14], CW [15] and ATMES [16]. It is noteworthy that the penalty function method needs to tune the punishment factor, and the MOEAs method brings additional objective. In this paper, CDP method is used to handle constraints, which requires no additional parameters.

The remainder of the paper is organized as follows. Section 2 designs a set of multi-objective constrained optimization problems (MCOPs). Section 3 introduces the repair operator. Section 4 gives the experimental results of the CTP and MCOP optimization problems, and Section 5 concludes the paper.

## 2. DESIGN OF MCOPs

The existing multi-objective constrained optimization problems mainly consist of CTP [22][23] and CF[24]. CTP benchmark problems can be defined as follows:

$$\begin{cases} \text{minimize } f_1(x) = x_1 \\ \text{minimize } f_2(x) = g(1 - \sqrt{f_1/g}) \\ g(x) = 1 + 9 \sum_{i=2}^{10}(x_i^2 - 10\cos(2\pi x_i) + 10) \\ \text{s.t. } C(x) \equiv \cos(\theta)(f_2(x) - e) - \sin(\theta)f_1(x) \geq \\ \quad a \mid \sin(b\pi(\sin(\theta)(f_2(x) - e) + \cos(\theta)f_1(x))^c) \mid^d \end{cases} \quad (2)$$

It is important to note that the problem can be made harder by setting $g(x)$ function with various local extreme. The inequality constraint C(x) has six parameters ($\theta$, a, b, c, d, and e). In fact, the above problem can be used as a constrained test problem generator by tuning these six parameters. Deb et al designed seven benchmark problems named CTP2-CTP8 by setting those six parameters. The original CTP2-CTP8 instances have only 2 decision variables and they are easy to solve. Hence, we extend the CTP2-CTP8 problems to ten decision variables and variable bounds are given by $0 \leq x_i \leq 1, i = 1,\ldots,10$. The six constraint parameters are the same as those used in [22].

According to the final report on CEC'09 MOEA competition, MOEA/D and NSGA-II are not quite suitable for solving CF instances. Even though it is very easy to search feasible solutions for CF, finding the true Pareto front turns out to be very difficult. This paper mainly focuses on applying the repair operators in the framework of MOEA/D and NSGA-II. Because CF is not a suitable test suite for MOEA/D and NSGA-II, we design a new set of multi-objective constrained optimization problems (MCOPs) to validate the proposed repair operator in the framework of MOEA/D and NSGA-II. Unlike CTP2-CTP8 instances which have the same multi-objectives and each problem has different constraint conditions by selecting six different parameters, we design a set of problems that have different multi-objectives but share the same constraint conditions. In terms of objective functions, we adopt ZDT test problems [28] but make some changes. In addition, nine ellipses are established in the objective space as the constraint conditions. The general form of constraint conditions are as follows:

$$C(x) \equiv ((f_1(x) - c_x) * \cos(\theta) - (f_2(x) - c_y) * \sin(\theta))^2 / a^2 \\ + ((f_1(x) - c_x) * \sin(\theta) + (f_2(x) - c_y) * \cos(\theta))^2 / b^2 \geq 1 \quad (3)$$

The constraint $C(x)$ has five parameters ($\theta, a, b, c_x$ and $c_y$), which can be used to further adjust the difficulty levels of the constraint conditions as needed. Among them $\theta$ denotes the counterclockwise rotation angle of the ellipse. $a$ and $b$ control the lengths of the long axis and minor axis of the ellipse respectively. $c_x$ and $c_y$ are two vectors representing the centers of the ellipses. For example, if we define the following parameters:

$$\begin{bmatrix} c_x \\ c_y \end{bmatrix} = \begin{bmatrix} 0 & 1 & 0 & 1 & 2 & 0 & 1 & 2 & 3 \\ 1.5 & 0.5 & 2.5 & 1.5 & 0.5 & 3.5 & 2.5 & 1.5 & 0.5 \end{bmatrix} \quad (4)$$

$$a^2 = 0.1, b^2 = 0.2, \theta = -0.25\pi$$

The distribution of constraining ellipses in the objective space is shown in Figure 1.

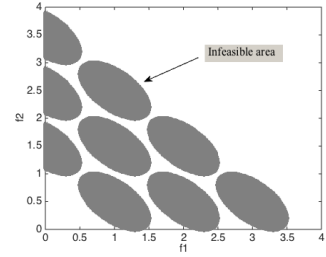

**Figure 1. The distribution of constraint functions.**

Combining the constraint functions with objective functions, we design seven multi-objective constraint optimization problems, namely MCOP1-MCOP7. The objective functions of them are listed in Table 1.

**Table 1. Objective functions of MCOP1-MCOP7.**

| Function Name | Function Definition |
|---|---|
| MCOP1 PF convex | $\begin{cases} \text{minimize } f_1(x) = g * x_1 \\ \text{minimize } f_2(x) = g(1 - \sqrt{f_1/g}) \\ g(x) = 1 + 9 \sum_{i=2}^{m} x_i / (m-1) \\ m = 30, x_i \in [0,1] \end{cases}$ |
| MCOP2 PF discrete | $\begin{cases} \text{minimize } f_1(x) = g * x_1 \\ \text{minimize } f_2(x) = g(1 - (f_1/g)^2) \\ g(x) = 1 + 9 \sum_{i=2}^{m} x_i / (m-1) \\ m = 30, x_i \in [0,1] \end{cases}$ |
| MCOP3<br>PF discrete | $\begin{cases} \text{minimize } f_1(x) = x_1 \\ \text{minimize } f_2(x) = g(1 - \sqrt{f_1/g}) \\ \quad - f_1 \sin(10\pi f_1) \\ g(x) = 1 + 9((\sum_{i=2}^{m} x_i / (m-1))^{0.25}) \\ m = 10, x_i \in [0,1] \end{cases}$ |
| MCOP4<br>PF convex | $\begin{cases} \text{minimize } f_1(x) = g * x_1 \\ \text{minimize } f_2(x) = g(1 - \sqrt{f_1/g}) \\ g(x) = 1 + 10(m-1) + \sum_{i=2}^{m} (x_i^2 - 10\cos(4\pi x_i)) \\ m = 10, x_i \in [0,1] \end{cases}$ |
| MCOP5<br>PF discrete | $\begin{cases} \text{minimize } f_1(x) = g * x_1 \\ \text{minimize } f_2(x) = g(1 - (f_1/g)^2) \\ g(x) = 1 + 10(m-1) + \sum_{i=2}^{m} (x_i^2 - 10\cos(4\pi x_i)) \\ m = 10, x_i \in [0,1] \end{cases}$ |
| MCOP6<br>PF discrete | $\begin{cases} \text{minimize } f_1(x) = 1 - \exp(-4x_1)\sin^6(6\pi x_1) \\ \text{minimize } f_2(x) = g(1 - (f_1/g)^2) \\ g(x) = 1 + 10(m-1) + \sum_{i=2}^{m} (x_i^2 - 10\cos(4\pi x_i)) \\ m = 10, x_i \in [0,1] \end{cases}$ |
| MCOP7<br>PF convex | $\begin{cases} \text{minimize } f_1(x) = 1 - \exp(-4x_1)\sin^6(6\pi x_1) \\ \text{minimize } f_2(x) = g(1 - \sqrt{f_1/g}) \\ g(x) = 1 + 10(m-1) + \sum_{i=2}^{m} (x_i^2 - 10\cos(4\pi x_i)) \\ m = 10, x_i \in [0,1] \end{cases}$ |

## 3. REPAIR OPERATOR

Repair operators are used to fix the infeasible solutions that violate the box constraints. A lot of research concentrates on repairing the infeasible solutions for discrete multi-objective constrained optimization problems. However, very few researchers have paid attention to the repair operators for continuous multi-objective constrained optimization problems. At present, there are two commonly used repair operators. One of the most commonly used repair operator can be defined as follows:

$$x_{i,j} = \begin{cases} L_j, \text{if } x_{i,j} < L_j \\ U_j, \text{if } x_{i,j} > U_j \end{cases} \quad (5)$$

Where $x_{i,j}$ represents the value of *j-th* component of individual *i*. $L_j$ denotes the lower bound of *j-th* component of the decision variables. $U_j$ denotes the upper bound of *j-th* component of the decision variables.

Another commonly used repair operator proposed by Wang etc. [26] can be defined as follows:

$$x_{i,j} = \begin{cases} \min\{U_j, 2L_j - x_{i,j}\}, \text{if } x_{i,j} < L_j \\ \max\{L_j, 2U_j - x_{i,j}\}, \text{if } x_{i,j} > U_j \end{cases} \quad (6)$$

In order to facilitate discussion, we denote the formula (5) and formula (6) as Repair-A and Repair-B respectively. In this paper, we propose a new repair operator denoted as Repair-C. The formula of our proposed repair operator can be defined as follows:

$$x_{i,j} = \begin{cases} U_j, \text{if } x_{i,j} < L_j \\ L_j, \text{if } x_{i,j} > U_j \end{cases} \quad (7)$$

This repair operator is inspired in part by the concept of opposition-based learning (OBL) originally introduced by Tizhoosh [29]. The main idea of OBL is, for finding a better candidate solution, simultaneous consideration of an estimate and its corresponding opposite estimate has a potential to help search towards the global optimum in a more efficient way, due to an arguably better preservation of diversity in the searching population. For example, the differential evolution process can be defined as follows:

$$x_{i,j}' = x_{i,j} + F(x_{r1,j} - x_{r2,j}) \quad (8)$$

Where $r_1$ and $r_2$ are two unequal random integers and not equal to $i$. $F$ denotes the factor of differential evolution, here we set $F = 0.5$. If $x_{i,j}'$ is less than its lower bound $L_j$, it can be inferred that $x_{i,j}$ has a higher probability of getting a value close to its lower bound $L_j$. In this case if we fix $x_{i,j}'$ to its upper bound $U_j$, which can be approximately considered as an opposite estimate of $x_{i,j}$, then this operator has a potential to increase the diversity of the population according to the philosophy of OBL. Even though this choice is a bit counterintuitive because normally people think fixing $x_{i,j}'$ to its lower bound $L_j$ is a better choice, but we shall also not ignore the possibility that fixing $x_{i,j}'$ to its lower bound $L_j$, which is a value with a lot loss of potential after many previous search attempts, the search may have a higher likelihood to be stuck in local minima. To verify this hypothesis, we conduct a lot of experiments which are described in detail in the Section of Experimental Study.

## 4. EXPERIMENTAL STUDY
### 4.1 Experimental Settings

In order to evaluate the performance of repair operators mentioned in section 3, we combined these three repair operators with NSGA-II and MOEA/D and then studied the experimental results on CTP2-CTP8 and MCOP1-MCOP7. Thirty independent runs with the six algorithms are conducted. The detailed parameter settings of these six algorithms are summarized as follows.

1) Setting for reproduction operators: The mutation probability $Pm = 1/n$ ($n$ is the number of decision variables) and its distribution index is set to be 20. For the DE operator, we set $CR = 1.0$ and $F = 0.5$ as recommended in [27].

2) Population size: $N = 200$.

3) Number of runs and stopping condition: Each algorithm runs 30 times independently on each test problems. The algorithm stops until 500 000 function evaluations.

4) Neighborhood size: $T = 20$.

5) Probability use to select in the neighborhood: $\delta = 0.9$.

6) The maximal number of solutions replaced by a child: $n_r = 2$.

## 4.2 Performance Metric

In this work, performance of a constrained multi-objective evolutionary algorithm is evaluated in two aspects – convergence and distribution. Convergence describes the closeness of the obtained Pareto front to the true Pareto front. Distribution on the other hand depicts how the solutions in the obtained Pareto are distributed. We select two metrics - inverted generation distance (IGD)[30] and hypervolume (HV)[30]. Detailed definitions of them are given as follows:

Inverted Generational Distance (IGD):

Let $p^*$ is the ideal Pareto front set, $A$ is an approximate Pareto front set achieved by evolutionary multi-objective algorithm. IGD metric denotes the distance between $p^*$ and $A$. It is defined as follows:

$$\begin{cases} IGD(P^*, A) = \dfrac{\sum_{y^* \in P^*} d(y^*, A)}{|P^*|} \\ d(y^*, A) = \min_{y \in A} \left\{ \sqrt{\sum_{i=1}^{m}(y_i^* - y_i)^2} \right\} \end{cases} \quad (8)$$

Where $m$ is the number of objectives, $|P^*|$ denotes the size of set $P^*$, $d(y^*, A)$ denotes the minimum Euclidean distance between $y^*$ and $A$. IGD metric can reflects the convergence and diversity simultaneously. The smaller IGD metric means the better performance.

Hypervolume (HV):

HV simultaneously considers the distribution of the obtained Pareto front $A$ and its vicinity to the true Pareto front. $HV$ is defined as the volume enclosed by $A$ and the reference vector $r = (r_1, r_2, \ldots, r_m)$. $HV$ can be defined as:

$$HV(P) = \bigcup_{i \in P} vol(i) \quad (9)$$

Here, $vol(i)$ represents the volume enclosed by solution $i \in A$ and the reference vector $r$. The maximum value of each objective in the ideal Pareto front set gives the value of each dimension of the reference point $r$, and thus constructs the reference point.

## 4.3 Experimental Result

In order to demonstrate the effectiveness of the proposed repair operator, we first compared it with the other two repair operators Operator_A and Operator_B (discussed in section III) in the framework of MOEA/D-CDP on CTP2-CTP8 and MCOP1-MCOP7 problems. The final populations with the best hypervolume metric in 30 independent runs with the three repair operators are shown in Figure 2.

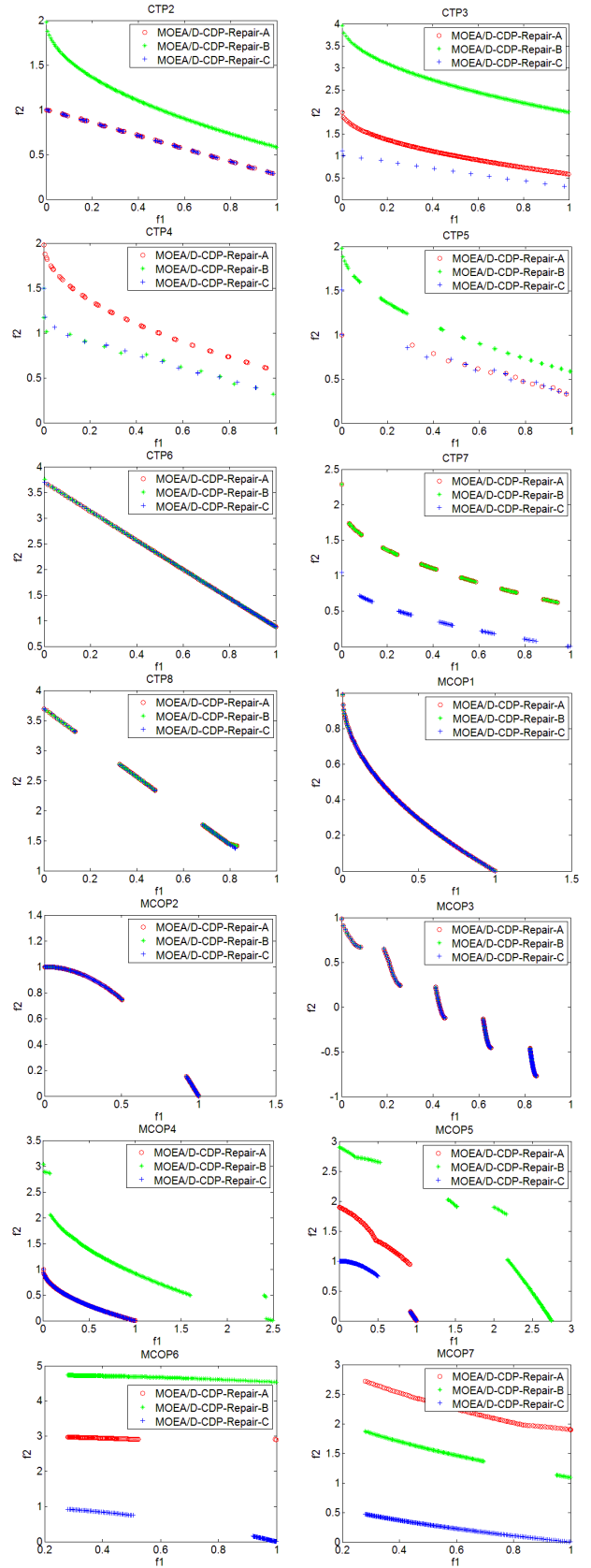

Figure 2. The final populations with the best hypervolume metric in 30 independent runs

From Figure 2, it is clear that MOEA/D-Repair-C has obtained best Pareto fronts on CTP3, CTP7, MCOP5, MCOP6 and MCOP7. For CTP2, CTP5 and MCOP4, MOEA/D-Repair-A and MOEA/D-Repair-C have a similar Pareto front, which is better than the Pareto front obtained by MOEA/D-Repair-B. MOEA/D-Repair-B and MOEA/D-Repair-C have a better Pareto front than MOEA/D-Repair-A on CTP4. For CTP6, CTP8, MCOP1, MCOP2 and MCOP3, the three methods have similar Pareto fronts. Overall MOEA/D-Repair-C outperforms or is at least competitive with MOEA/D-Repair-A and MOEA/D-Repair-B in all test cases.

**Table 2. IGD values of MOEA/D-Repair-A and MOEA/D-Repair-B**

| Instance | MOEA/D-Repair-A | | MOEA/D-Repair-B | |
|---|---|---|---|---|
| - | Mean | Std. | Mean | Std. |
| CTP2 | **6.81E-02** | 3.86E-02 | 1.46E-01 | 7.84E-02 |
| CTP3 | **1.63E-01** | 9.45E-02 | 3.37E-01 | 1.71E-01 |
| CTP4 | **4.68E-01** | 2.21E-01 | 5.88E-01 | 4.67E-01 |
| CTP5 | **8.08E-02** | 3.70E-02 | 1.38E-01 | 9.20E-02 |
| CTP6 | **1.29E-02** | 9.65E-03 | 1.28E-01 | 7.51E-02 |
| CTP7 | **1.17E-01** | 4.86E-02 | 1.58E-01 | 7.13E-02 |
| CTP8 | **3.77E-02** | 5.80E-02 | 2.72E-01 | 1.06E-01 |
| MCOP1 | **2.39E-04** | 2.33E-06 | 2.40E-04 | 3.84E-06 |
| MCOP2 | 2.56E-04 | 1.47E-07 | 2.56E-04 | 1.71E-07 |
| MCOP3 | 2.71E-04 | 4.94E-07 | 2.71E-04 | 1.82E-06 |
| MCOP4 | **8.96E-02** | 3.01E-02 | 1.21E-01 | 2.85E-02 |
| MCOP5 | **1.67E-01** | 6.14E-02 | 2.88E-01 | 9.92E-02 |
| MCOP6 | **1.03E-01** | 2.05E-02 | 1.23E-01 | 2.47E-02 |
| MCOP7 | **1.09E-01** | 3.09E-02 | 1.28E-01 | 3.80E-02 |

**Table 3. IGD values of MOEA/D-Repair-A and MOEA/D-Repair-C**

| Instance | MOEA/D-Repair-A | | MOEA/D-Repair-C | |
|---|---|---|---|---|
| - | Mean | Std. | Mean | Std. |
| CTP2 | 6.81E-02 | 3.86E-02 | **1.70E-04** | 3.15E-06 |
| CTP3 | 1.63E-01 | 9.45E-02 | **1.40E-03** | 1.71E-04 |
| CTP4 | 4.68E-01 | 2.21E-01 | **4.41E-02** | 2.83E-02 |
| CTP5 | 8.08E-02 | 3.70E-02 | **6.85E-03** | 3.88E-03 |
| CTP6 | 1.29E-02 | 9.65E-03 | **5.04E-04** | 1.99E-05 |
| CTP7 | 1.17E-01 | 4.86E-02 | **1.39E-04** | 2.77E-07 |
| CTP8 | 3.77E-02 | 5.80E-02 | **1.12E-03** | 1.32E-04 |
| MCOP1 | 2.39E-04 | 2.33E-06 | **2.37E-04** | 3.68E-06 |
| MCOP2 | 2.56E-04 | 1.47E-07 | 2.56E-04 | 4.65E-07 |
| MCOP3 | 2.71E-04 | 4.94E-07 | 2.71E-04 | 3.03E-06 |
| MCOP4 | 8.96E-02 | 3.01E-02 | **1.55E-02** | 1.62E-02 |
| MCOP5 | 1.67E-01 | 6.14E-02 | **3.27E-02** | 3.49E-02 |
| MCOP6 | 1.03E-01 | 2.05E-02 | **2.89E-02** | 2.08E-02 |
| MCOP7 | 1.09E-01 | 3.09E-02 | **4.63E-02** | 1.60E-02 |

**Table 4. T-test values of IGD among MOEA/D-Repair-A, MOEA/D-Repair-B and MOEA/D-Repair-C**

| Instance | Repair-C vs Repair-A | | Repair-C vs Repair-B | |
|---|---|---|---|---|
| - | h-value | p-value | h-value | p-value |
| CTP2 | **1.00E+00** | 5.66E-14 | **1.00E+00** | 8.11E-15 |
| CTP3 | **1.00E+00** | 1.53E-13 | **1.00E+00** | 8.79E-16 |
| CTP4 | **1.00E+00** | 3.64E-15 | **1.00E+00** | 1.67E-08 |
| CTP5 | **1.00E+00** | 6.17E-16 | **1.00E+00** | 7.15E-11 |
| CTP6 | **1.00E+00** | 1.20E-09 | **1.00E+00** | 2.41E-13 |
| CTP7 | **1.00E+00** | 2.36E-19 | **1.00E+00** | 8.56E-18 |
| CTP8 | **1.00E+00** | 5.13E-04 | **1.00E+00** | 1.47E-20 |
| MCOP1 | **1.00E+00** | 1.91E-02 | **1.00E+00** | 1.62E-03 |
| MCOP2 | 0.00E+00 | 9.80E-01 | 0.00E+00 | 9.82E-01 |
| MCOP3 | 0.00E+00 | 8.17E-01 | 0.00E+00 | 8.92E-01 |
| MCOP4 | **1.00E+00** | 1.78E-17 | **1.00E+00** | 2.37E-25 |
| MCOP5 | **1.00E+00** | 3.71E-15 | **1.00E+00** | 1.55E-19 |
| MCOP6 | **1.00E+00** | 2.42E-20 | **1.00E+00** | 4.34E-23 |
| MCOP7 | **1.00E+00** | 2.41E-14 | **1.00E+00** | 5.67E-16 |

Table 2 and Table 3 present the average values of IGD over 30 independent runs in the framework of MOEA/D. Table 4 presents the t-test values of IGD among three different repair operators. It can be observed that MOEA/D-Repair-C performs significantly better than other two methods on all the instances except for MCOP2 and MCOP3, and almost the same as the other two kinds of repair operators on MCOP2 and MCOP3 (with slightly bigger standard deviation). The main cause is that MCOP2 and MCOP3 (actually including MCOP1) are relatively easy to solve that the three different repair operators can not reflect differences on them.

**Table 5. HV values of MOEA/D-Repair-A and MOEA/D-Repair-B**

| HV | MOEA/D-Repair-A | | MOEA/D-Repair-B | |
|---|---|---|---|---|
| - | Mean | Std. | Mean | Std. |
| CTP2 | **5.00E-02** | 1.02E-01 | 9.76E-03 | 3.71E-02 |
| CTP3 | **3.83E-02** | 6.45E-02 | 0.00E+00 | 0.00E+00 |
| CTP4 | 8.43E-03 | 3.21E-02 | **3.84E-02** | 8.43E-02 |
| CTP5 | **2.16E-02** | 7.48E-02 | 1.83E-02 | 4.75E-02 |
| CTP6 | **4.04E-01** | 8.34E-02 | 1.14E-01 | 1.89E-01 |
| CTP7 | **2.43E-03** | 5.53E-03 | 9.74E-04 | 3.71E-03 |
| CTP8 | **3.08E-01** | 1.40E-01 | 5.59E-02 | 1.46E-01 |
| MCOP1 | 6.64E-01 | 1.78E-05 | 6.64E-01 | 2.05E-05 |
| MCOP2 | 2.21E-01 | 8.54E-06 | 2.21E-01 | 1.50E-05 |
| MCOP3 | **5.16E-01** | 3.43E-06 | 5.15E-01 | 2.23E-05 |
| MCOP4 | **3.37E-02** | 1.22E-01 | 2.53E-04 | 9.64E-04 |
| MCOP5 | **3.55E-03** | 1.40E-02 | 0.00E+00 | 0.00E+00 |
| MCOP6 | 0.00E+00 | 0.00E+00 | 0.00E+00 | 0.00E+00 |
| MCOP7 | 0.00E+00 | 0.00E+00 | 0.00E+00 | 0.00E+00 |

**Table 6. HV values of MOEA/D-Repair-A and MOEA/D-Repair-C**

| Instance | MOEA/D-Repair-A | | MOEA/D-Repair-C | |
|---|---|---|---|---|
| - | Mean | Std. | Mean | Std. |
| CTP2 | 5.00E-02 | 1.02E-01 | **4.77E-01** | 9.83E-05 |
| CTP3 | 3.83E-02 | 6.45E-02 | **4.45E-01** | 1.90E-03 |
| CTP4 | 8.43E-03 | 3.21E-02 | **3.08E-01** | 8.97E-02 |
| CTP5 | 2.16E-02 | 7.48E-02 | **2.52E-01** | 1.03E-01 |
| CTP6 | 4.04E-01 | 8.34E-02 | **4.99E-01** | 2.21E-04 |
| CTP7 | 2.43E-03 | 5.53E-03 | **5.46E-01** | 2.56E-05 |
| CTP8 | 3.08E-01 | 1.40E-01 | **4.41E-01** | 4.72E-04 |
| MCOP1 | 6.64E-01 | 1.78E-05 | 6.64E-01 | 6.06E-05 |
| MCOP2 | 2.21E-01 | 8.54E-06 | 2.21E-01 | 4.60E-05 |
| MCOP3 | **5.16E-01** | 3.43E-06 | 5.15E-01 | 2.98E-05 |
| MCOP4 | 3.37E-02 | 1.22E-01 | **4.93E-01** | 1.70E-01 |
| MCOP5 | 3.55E-03 | 1.40E-02 | **1.14E-01** | 1.09E-01 |
| MCOP6 | 0.00E+00 | 0.00E+00 | **5.90E-02** | 1.02E-01 |
| MCOP7 | 0.00E+00 | 0.00E+00 | **4.22E-02** | 1.20E-01 |

**Table 7. T-test values of HV among MOEA/D-Repair-A, MOEA/D-Repair-B and MOEA/D-Repair-C**

| Instance | Repair-C vs Repair-A | | Repair-C vs Repair-B | |
|---|---|---|---|---|
| - | h-value | p-value | h-value | p-value |
| CTP2 | **1.00E+00** | 4.89E-31 | **1.00E+00** | 1.23E-57 |
| CTP3 | **1.00E+00** | 1.12E-40 | **1.00E+00** | 4.07E-131 |
| CTP4 | **1.00E+00** | 9.40E-25 | **1.00E+00** | 1.26E-17 |
| CTP5 | **1.00E+00** | 2.21E-14 | **1.00E+00** | 1.59E-16 |
| CTP6 | **1.00E+00** | 3.21E-08 | **1.00E+00** | 2.39E-16 |
| CTP7 | **1.00E+00** | 2.77E-109 | **1.00E+00** | 1.97E-119 |
| CTP8 | **1.00E+00** | 1.16E-06 | **1.00E+00** | 3.15E-21 |
| MCOP1 | 0.00E+00 | 1.00E+00 | 0.00E+00 | 1.00E+00 |
| MCOP2 | 0.00E+00 | 1.00E+00 | 0.00E+00 | 1.00E+00 |
| MCOP3 | 0.00E+00 | 1.00E+00 | 0.00E+00 | 1.00E+00 |
| MCOP4 | **1.00E+00** | 1.12E-17 | **1.00E+00** | 4.46E-23 |
| MCOP5 | **1.00E+00** | 3.85E-07 | **1.00E+00** | 1.65E-07 |
| MCOP6 | **1.00E+00** | 1.19E-03 | **1.00E+00** | 1.19E-03 |
| MCOP7 | **1.00E+00** | 2.97E-02 | **1.00E+00** | 2.97E-02 |

Table 5 and Table 6 present the average values of HV over 30 independent runs in the framework of MOEA/D. Table 7 presents the t-test values of HV among three repair operators. From Table 5, Table 6 and Table 7, it can be observed that MOEA/D-Repair-C performs significantly better than the other two kinds of methods on all the instances except for MCOP1, MCOP2 and MCOP3, which means that our proposed repair operator can effectively avoid the population falling into local optimum in the framework of MOEA/D. For MCOP1, MCOP2 and MCOP3, the three methods obtain almost the same results.

**Table 8. IGD values of NSGAII-Repair-A and NSGAII-Repair-B**

| Instance | NSGAII-Repair-A | | NSGAII-Repair-B | |
|---|---|---|---|---|
| - | Mean | Std. | Mean | Std. |
| CTP2 | **4.66E-02** | 3.23E-02 | 1.17E-01 | 6.41E-02 |
| CTP3 | **1.03E-01** | 6.32E-02 | 3.12E-01 | 1.45E-01 |
| CTP4 | **2.90E-01** | 1.67E-01 | 6.18E-01 | 3.26E-01 |
| CTP5 | **6.14E-02** | 2.32E-02 | 1.12E-01 | 5.13E-02 |
| CTP6 | **1.08E-02** | 7.22E-03 | 8.48E-02 | 7.72E-02 |
| CTP7 | **7.86E-02** | 4.44E-02 | 1.61E-01 | 8.43E-02 |
| CTP8 | **1.98E-02** | 9.15E-03 | 1.61E-01 | 1.47E-01 |
| MCOP1 | **2.74E-04** | 1.83E-05 | 4.52E-04 | 4.21E-05 |
| MCOP2 | **1.07E-02** | 2.04E-02 | 1.60E-02 | 1.05E-02 |
| MCOP3 | **1.18E-04** | 4.38E-06 | 2.57E-04 | 4.12E-05 |
| MCOP4 | **1.01E-01** | 3.14E-02 | 1.36E-01 | 6.00E-02 |
| MCOP5 | **2.27E-01** | 8.64E-02 | 3.36E-01 | 1.05E-01 |
| MCOP6 | **1.02E-01** | 4.07E-02 | 1.57E-01 | 5.35E-02 |
| MCOP7 | **1.13E-01** | 3.51E-02 | 1.70E-01 | 4.87E-02 |

**Table 9. IGD values of NSGAII-Repair-A and NSGAII-Repair-C**

| Instance | NSGAII-Repair-A | | NSGAII-Repair-C | |
|---|---|---|---|---|
| - | Mean | Std. | Mean | Std. |
| CTP2 | 4.66E-02 | 3.23E-02 | **1.13E-04** | 4.49E-05 |
| CTP3 | 1.03E-01 | 6.32E-02 | **2.95E-03** | 9.83E-04 |
| CTP4 | 2.90E-01 | 1.67E-01 | **7.46E-02** | 2.32E-02 |
| CTP5 | 6.14E-02 | 2.32E-02 | **1.01E-02** | 4.84E-03 |
| CTP6 | 1.08E-02 | 7.22E-03 | **3.02E-04** | 1.16E-04 |
| CTP7 | 7.86E-02 | 4.44E-02 | **5.25E-05** | 1.54E-06 |
| CTP8 | 1.98E-02 | 9.15E-03 | **5.05E-04** | 8.50E-05 |
| MCOP1 | **2.74E-04** | 1.83E-05 | 7.20E-03 | 1.61E-02 |
| MCOP2 | **1.07E-02** | 2.04E-02 | 2.56E-02 | 2.14E-02 |
| MCOP3 | **1.18E-04** | 4.38E-06 | 1.45E-02 | 1.80E-02 |
| MCOP4 | 1.01E-01 | 3.14E-02 | **2.72E-04** | 1.75E-05 |
| MCOP5 | 2.27E-01 | 8.64E-02 | **1.83E-04** | 1.54E-05 |
| MCOP6 | 1.02E-01 | 4.07E-02 | **4.31E-05** | 1.03E-05 |
| MCOP7 | 1.13E-01 | 3.51E-02 | **9.72E-05** | 1.92E-05 |

**Table 10. T-test values of IGD among NSGAII-Repair-A, NSGAII-Repair-B and NSGAII-Repair-C**

| Instance | Repair-C vs Repair-A | | Repair-C vs Repair-B | |
|---|---|---|---|---|
| - | h-value | p-value | h-value | p-value |
| CTP2 | **1.00E+00** | 4.90E-11 | **1.00E+00** | 1.89E-14 |
| CTP3 | **1.00E+00** | 2.07E-12 | **1.00E+00** | 3.82E-17 |
| CTP4 | **1.00E+00** | 1.39E-09 | **1.00E+00** | 4.31E-13 |
| CTP5 | **1.00E+00** | 2.00E-17 | **1.00E+00** | 6.34E-16 |
| CTP6 | **1.00E+00** | 3.86E-11 | **1.00E+00** | 6.92E-08 |
| CTP7 | **1.00E+00** | 5.03E-14 | **1.00E+00** | 2.61E-15 |
| CTP8 | **1.00E+00** | 5.44E-17 | **1.00E+00** | 8.16E-08 |
| MCOP1 | 0.00E+00 | 9.89E-01 | 0.00E+00 | 9.87E-01 |
| MCOP2 | 0.00E+00 | 9.96E-01 | 0.00E+00 | 9.84E-01 |
| MCOP3 | 0.00E+00 | 1.00E+00 | 0.00E+00 | 1.00E+00 |
| MCOP4 | **1.00E+00** | 3.48E-25 | **1.00E+00** | 2.90E-18 |
| MCOP5 | **1.00E+00** | 4.03E-21 | **1.00E+00** | 3.36E-25 |
| MCOP6 | **1.00E+00** | 4.30E-20 | **1.00E+00** | 2.54E-23 |
| MCOP7 | **1.00E+00** | 2.60E-25 | **1.00E+00** | 5.48E-27 |

Table 8 and table 9 present the average values of IGD over 30 independent runs in the framework of NSGA-II. Table 10 presents the t-test values of IGD among the three repair operators. It can be observed that similar to the results obtained in the framework of MOEA/D, the Repair-C performs better than Repair-A and Repair-B on all the instances except for MCOP1, MCOP2 and MCOP3. For MCOP1, MCOP2 and MCOP3, the three methods obtain almost the same results.

**Table 11. HV values of NSGAII-Repair-A and NSGAII-Repair-B**

| Instance | NSGAII-Repair-A | | NSGAII-Repair-B | |
| --- | --- | --- | --- | --- |
| - | Mean | Std. | Mean | Std. |
| CTP2 | **4.77E-01** | 1.01E-01 | 2.42E-02 | 5.49E-02 |
| CTP3 | **4.37E-01** | 7.15E-02 | 9.47E-03 | 3.61E-02 |
| CTP4 | **3.43E-01** | 4.51E-02 | 8.42E-03 | 3.21E-02 |
| CTP5 | **1.37E-01** | 2.29E-02 | 1.37E-02 | 4.19E-02 |
| CTP6 | **5.00E-01** | 4.24E-01 | 1.82E-01 | 2.02E-01 |
| CTP7 | **5.47E-01** | 4.14E-02 | 1.97E-02 | 9.96E-02 |
| CTP8 | **4.45E-01** | 3.63E-01 | 1.80E-01 | 1.94E-01 |
| MCOP1 | 6.64E-01 | 6.64E-01 | **6.59E-01** | 6.29E-04 |
| MCOP2 | **2.21E-01** | 1.92E-01 | 1.27E-01 | 5.24E-02 |
| MCOP3 | **5.17E-01** | 5.17E-01 | 5.10E-01 | 1.57E-03 |
| MCOP4 | **2.63E-02** | 8.14E-03 | 1.15E-02 | 4.91E-02 |
| MCOP5 | **2.21E-01** | 7.38E-03 | 0.00E+00 | 0.00E+00 |
| MCOP6 | **2.38E-01** | 2.38E-02 | 0.00E+00 | 0.00E+00 |
| MCOP7 | **5.48E-01** | 1.83E-02 | 0.00E+00 | 0.00E+00 |

**Table12. HV values of NSGAII-Repair-A and NSGAII-Repair-C**

| Instance | NSGAII-Repair-a | | NSGAII-Repair-c | |
| --- | --- | --- | --- | --- |
| - | Mean | Std. | Mean | Std. |
| CTP2 | 1.01E-01 | 1.45E-01 | **4.77E-01** | 5.21E-04 |
| CTP3 | 7.15E-02 | 9.84E-02 | **4.31E-01** | 9.21E-03 |
| CTP4 | 4.51E-02 | 7.97E-02 | **1.50E-01** | 5.88E-02 |
| CTP5 | 2.29E-02 | 5.21E-02 | **1.54E-01** | 6.62E-02 |
| CTP6 | 4.24E-01 | 6.14E-02 | **5.00E-01** | 8.20E-04 |
| CTP7 | 4.14E-02 | 1.38E-01 | **5.47E-01** | 4.14E-05 |
| CTP8 | 3.63E-01 | 3.61E-02 | **4.44E-01** | 2.12E-04 |
| MCOP1 | **6.64E-01** | 5.97E-05 | 5.85E-01 | 1.78E-01 |
| MCOP2 | **1.92E-01** | 3.45E-02 | 8.03E-02 | 8.81E-02 |
| MCOP3 | **5.17E-01** | 2.29E-05 | 3.15E-01 | 2.08E-01 |
| MCOP4 | 8.14E-03 | 1.21E-02 | **6.64E-01** | 1.91E-04 |
| MCOP5 | 7.38E-03 | 4.04E-02 | **2.21E-01** | 4.78E-05 |
| MCOP6 | 2.38E-02 | 7.26E-02 | **2.38E-01** | 1.55E-05 |
| MCOP7 | 1.83E-02 | 1.00E-01 | **5.48E-01** | 1.86E-04 |

**Table 13. T-test values of HV among NSGAII-Repair-A, NSGAII-Repair-B and NSGAII-Repair-C**

| Instance | Repair-C vs Repair-A | | Repair-C vs Repair-B | |
| --- | --- | --- | --- | --- |
| - | h-value | p-value | h-value | p-value |
| CTP2 | **1.00E+00** | 7.84E-21 | **1.00E+00** | 3.56E-47 |
| CTP3 | **1.00E+00** | 6.17E-28 | **1.00E+00** | 4.80E-55 |
| CTP4 | **1.00E+00** | 1.52E-07 | **1.00E+00** | 5.41E-17 |
| CTP5 | **1.00E+00** | 3.72E-12 | **1.00E+00** | 2.87E-14 |
| CTP6 | **1.00E+00** | 2.94E-09 | **1.00E+00** | 2.69E-12 |
| CTP7 | **1.00E+00** | 3.84E-28 | **1.00E+00** | 1.67E-36 |
| CTP8 | **1.00E+00** | 4.47E-18 | **1.00E+00** | 2.47E-10 |
| MCOP1 | 0.00E+00 | 9.91E-01 | 0.00E+00 | 9.87E-01 |
| MCOP2 | 0.00E+00 | 1.00E+00 | 0.00E+00 | 9.93E-01 |
| MCOP3 | 0.00E+00 | 1.00E+00 | 0.00E+00 | 1.00E+00 |
| MCOP4 | **1.00E+00** | 2.43E-94 | **1.00E+00** | 5.43E-59 |
| MCOP5 | **1.00E+00** | 1.64E-36 | **1.00E+00** | 2.72E-206 |
| MCOP6 | **1.00E+00** | 1.95E-23 | **1.00E+00** | 1.81E-236 |
| MCOP7 | **1.00E+00** | 1.67E-36 | **1.00E+00** | 6.88E-195 |

Table 11 and Table 12 present the average values of HV over 30 independent runs in the framework of NSGA-II. Table 13 presents the t-test values of HV among three repair operators. From Table 11, Table 12 and Table 13, it can be observed that NSGA-II-Repair-C performs significantly better than other two kinds of methods on all the instances except for MCOP1, MCOP2 and MCOP3. For MCOP1, MCOP2 and MCOP3, the three methods obtain almost the same results. It can be therefore concluded that the proposed repair operator can also work well in the framework of NSGA-II.

## 5. CONCLUSION

This paper proposes a new repair operator which employs a reversed correction strategy to fix the solutions that violate the box-constraint. In order to validate its performance on convergence and diversity, a new set of constrained multi-objective optimization problems is designed, to complement the well-known CTP test suite. The performance of the proposed repair operator has been compared with the other two kinds of commonly used repair operators. Experimental results show that it outperforms the other repair operators in terms of convergence and diversity, based on the two classic frameworks of MOEA/D and NSGA-II. The future work includes combining the proposed repair operator with other state-of-the-art algorithms to further validate the repair operator and improve the performance of the algorithms, and testing them in real-world applications.